\documentclass[10pt,twocolumn,letterpaper]{article}

\usepackage[pagenumbers]{cvpr} 
\usepackage[english]{babel}
\usepackage{graphicx}
\usepackage{amsmath}
\usepackage{amssymb}
\usepackage{booktabs}

\usepackage[pagebackref,breaklinks,colorlinks]{hyperref}

\usepackage[capitalize]{cleveref}
\crefname{section}{Sec.}{Secs.}
\Crefname{section}{Section}{Sections}
\Crefname{table}{Table}{Tables}
\crefname{table}{Tab.}{Tabs.}

\def\cvprPaperID{*****} 
\def\confName{CVPR}
\def\confYear{2022}

\begin{document}

\title{Applying Spatiotemporal Attention to Identify Distracted and Drowsy Driving with Vision Transformers}

\author{Samay Lakhani\\
Jericho High School\\
99 Cedar Swamp Rd, Jericho, NY 11753\\
{\tt\small samay.lakhani@gmail.com}
}
\maketitle

\begin{abstract}
   A 20\% rise in car crashes in 2021 compared to 2020 has been observed as a result of increased distraction and drowsiness. Drowsy and distracted driving are the cause of 45\% of all car crashes. As a means to decrease drowsy and distracted driving, detection methods using computer vision can be designed to be low-cost, accurate, and minimally invasive. This work investigated the use of the vision transformer to outperform state-of-the-art accuracy from 3D-CNNs. Two separate transformers were trained for drowsiness and distractedness. The drowsy video transformer model was trained on the National Tsing-Hua University Drowsy Driving Dataset (NTHU-DDD) with a Video Swin Transformer model for 10 epochs on two classes - drowsy and non-drowsy simulated over 10.5 hours. The distracted video transformer was trained on the Driver Monitoring Dataset (DMD) with Video Swin Transformer for 50 epochs over 9 distraction-related classes. The accuracy of the drowsiness model reached 44\% and a high loss value on the test set, indicating overfitting and poor model performance. Overfitting indicates limited training data and applied model architecture lacked quantifiable parameters to learn. The distracted model outperformed state-of-the-art models on DMD reaching 97.5\%, indicating that with sufficient data and a strong architecture, transformers are suitable for unfit driving detection. Future research should use newer and stronger models such as TokenLearner to achieve higher accuracy and efficiency, merge existing datasets to expand to detecting drunk driving and road rage to create a comprehensive solution to prevent traffic crashes, and deploying a functioning prototype to revolutionize the automotive safety industry.
\end{abstract}

\section{Introduction}
\label{sec:intro}

In 2021, there was a 20\% increase in traffic crashes compared to 2020\cite{Early_Estimate_of_Motor_Vehicle_Traffic_Fatalities_for_the_First_Half}. Following the onset of the COVID-19 pandemic, the lines between work, school, and home have been blurred. Increased reliance on electronic systems for communications has created a heavy impact on driving crashes and fatalities.

Possible solutions to detecting unfit driving have been implemented including applying EEGs, alcohol monitors, and lane detection systems. However, these are either invasive, expensive, or not scalable to detect all kinds of unfit driving, or some combination of the three\cite{Mardi_Ashtiani_Mikaili_2011,Li_Downen_Dong_Tran_LeSaux_Meltzer_Li_2019, Riera_Ozcan_Merickel_Rizzo_Sarkar_Sharma_2019}. For example, an alcohol monitor cannot identify if the driver is drowsy. Another solution that has gained popularity is computer vision, which is increasingly used by car manufacturers such as Toyota, Subaru, and Tesla \cite{Twitter_Instagram_Email_Facebook_2021}.

Computer vision involves monitoring the driver’s movements and analyzing the vision data with machine learning. The two main approaches can be categorized into explicit and implicit methods that monitor specific human features like PERCLOS, EAR, and human pose tracking \cite{George_Routray_2015,Yusri_Mangat_Wasenmuller_2021, 9548625}. 

Since all the features are hand-engineered, implicit methods reach high accuracy, but lack scalability to more than one category of unfit driving (i.e. drowsy or distracted driving). Implicit methods involve 3D CNNs and video classifiers. Video classifiers implicitly learn the features involved with each category of unfit driving. This method can scale to several categories (i.e. drowsy and distracted), but investigations due to a lack of data that spans several categories. Implicit methods lack accuracy compared to explicit methods\cite{Wijnands_Thompson_Nice_Aschwanden_Stevenson_2020}. Since each method exclusively is scalable or accurate, there are no comprehensive solutions to solving unfit driving yet. A new architecture, the vision transformer, may serve as a solution.
\begin{equation} \label{eq:1}
Attention(Q, K, V) = softmax(\frac{QK^T}{\sqrt{d_k}})V
\end{equation}

Introduced in 2017, the transformer mechanism is revolutionizing natural language processing and computer vision \cite{Vaswani_Shazeer_Parmar_Uszkoreit_Jones_Gomez_Kaiser_Polosukhin_2017}. Transformers rely on “attention,” (Equation \ref{eq:1}) a mechanism for assigning different importance to segments of the input data. In other words, it “pays attention” to the most important part of the data. The transformer relies on the Query ($Q$), Key ($K$), and Value ($V$) vectors to compute attention.

Transformers are now the model of choice for high-parameter language models such as BERT, GPT-3, Megatron-Turing NLG for their scalability of parameters with respect to performance \cite{Devlin_Chang_Lee_Toutanova_2019,BrownMann2020,Shoeybi_Patwary_Puri_LeGresley_Casper_Catanzaro_2020}.
Following success in NLP, transformers were applied to computer vision. Applying transformers to vision tasks is a challenging task as images contain a greater amount of raw data in pixels compared to language tasks, requiring scaling of the transformer. The first major success with transformers in vision tasks was ViT, which – instead of applying attention to pixels – applied attention to non-overlapping patches of the image \cite{Dosovitskiy2021}. This method reached competitive accuracy with CNNs on ImageNet.

Convolutional neural networks (CNNs) suffered from diminishing returns with an increase in parameters, and new works built on non-overlapping methods \cite{Boesch_2022}, implementing more complex processing operations. A notable example is Swin Transformer. Swin Transformer relies on shifted windows for a more rich understanding of the data\cite{Liu_Lin_Cao_Hu_Wei_Zhang_Lin_Guo_2021}. Video Swin Transformer expanded this operation to videos with tubelet embeddings and using 3D windows. Swin and its video counterpart – Video Swin Transformer – reach among the top accuracies for computer vision tasks for ImageNet and Kinetics-400, respectively, outperforming ViT\cite{Liu_Ning_Cao_Wei_Zhang_Lin_Hu_2021}.
Although vision transformers have promising results for computer vision, they have never been applied for unfit driving. This experiment details a) the application of two transformer architectures applied to drowsy and distracted driving to understand if vision transformers can compete – or outperform – 3D CNNs to create a comprehensive and accurate solution to unfit driving and b) an efficient transformer for implementation in a mobile device or embedded system, like a Jetson Nano.

\begin{figure}
  \centering
  \includegraphics[width=0.8\linewidth]{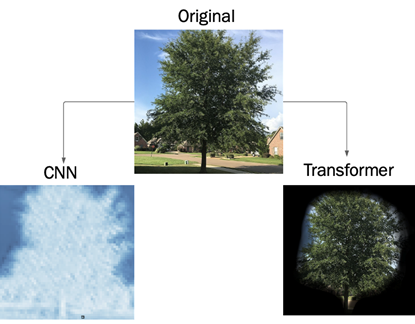}

  \caption{
  Difference in processing methods for convolutional neural networks and transformer. CNNs extract edges while the transformer extracts the most important spatial information from the input data. }
  \label{fig:onecol1}
\end{figure}

\section{Methodology}

Two experiments were performed to evaluate the accuracy of vision transformers for distracted and drowsy on pre-existing datasets. The first experiment focused on drowsy driving, and the second focused on distracted driving. The model tested in both experiments was a Video Swin Transformer architecture adapting the main features from \cite{Liu_Ning_Cao_Wei_Zhang_Lin_Hu_2021}. 

Efficiency was considered in model hyperparameters to achieve the second engineering goal of implementation in mobile devices.

\subsection{Experiment 1. Drowsy Driving}
For drowsy driving, a Video Swin Transformer architecture from \cite{Liu_Ning_Cao_Wei_Zhang_Lin_Hu_2021} was adapted for video classification and used to evaluate the performance of transformers for unfit driving. 

The input data was first passed to a CNN feature map extractor to extract a 1024-dimensional vector. The CNN feature extractor serves to reduce the computational cost by reducing the size of the input data. Then, the feature map is passed through an embedding layer that encodes per-pixel understandings. The embeddings are passed through the transformer encoder closely following \cite{Liu_Ning_Cao_Wei_Zhang_Lin_Hu_2021} transformer architecture. Next is 1D max pooling, a dense layer with 0.5 dropout, then a final output softmax layer for classification. 

The data used for the first experiment was the National Tsing Hua University Drowsy Driving Dataset (NTHU-DDD) \cite{}. NTHU-DDD is an existing and open benchmark video dataset for drowsy driving with 9.5 hours of footage in diverse environments. The footage was recorded of 18 subjects driving in combinations of glasses, sunglasses, dark, and light environments. A sequence length of 30 frames was extracted per training sample. An 80/20 train/test split was used. 

The model was trained on a Google Colaboratory notebook for 10 epochs with a Tesla K80 GPU.

\begin{figure*}[t]
  \centering
  \includegraphics[width=0.8\linewidth]{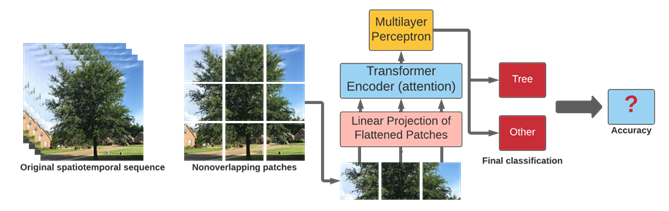}

  \caption{
Illustration of the vision transformer architecture adapted from  \cite{Dosovitskiy2021} inferencing on spatiotemporal data to output a prediction of drowsy/alert. The multi-head attention encoder in the transformer encoder allows for transformers to “pay attention” to only the most important features and reach higher accuracies than those of 3D CNNs.}
  \label{fig:onecol}
\end{figure*}
\subsection{Experiment 2. Distracted Driving}
For distracted driving, the Video Swin Transformer architecture was applied to understand the impact of a larger model size. Compared to Experiment I, Video Swin Transformer requires 22 more GFLOPS than ViT. Video Swin Transformer was used as it achieves higher accuracy than a vanilla transformer on benchmark datasets.

	The input data is tokenized with a tubelet embedding, extracting T (frames) x H (height) x W (width) x C (color channels) patches. Shifted windows are extracted from the video in 3-dimensional rectangular prisms from the video. The windows are shifted and sometimes overlap to create residual connections between spatial and temporal regions. These residual connections do not exist in ViT as all patches are extracted uniformly. Each 2x4x4x3 patch was tokenized to produce a 96-dimensional feature. To preserve temporal information, dimension T is not reduced, extracted, or compressed in any way. Patch merging is applied to reduce spatial information 2x, passed through layer normalization, and multi-head self-attention (MSA) is performed. 
	
The data used for the second experiment was the Driver Monitoring Dataset (DMD) \cite{Ortega_Kose_Canas_Chao_Unnervik_Nieto_Otaegui_Salgado_2020}. DMD is an ongoing dataset project, focusing on distraction-related categories such as texting, calling, and adjusting the radio. It is an existing and open benchmark video dataset for distracted driving with 40.75 hours of footage in diverse environments. The footage was recorded of 10 subjects driving in car simulators and closed parking lots. A sequence length of 30 frames was extracted per training sample, following methods from previous literature \cite{Wijnands_Thompson_Nice_Aschwanden_Stevenson_2020}. An 80/20 train/test split was used. 

	The model was trained with a Tesla T4 GPU considering the higher computational cost compared to Experiment I because of the larger architecture.
	
The accuracy of the model is defined by the number of correct predictions divided by the number of total predictions. Once trained, the accuracy of the models on the testing dataset was compared to previous literature with 3D CNNs for drowsy driving and distracted driving which currently reach 75.4\% and 97.2\% respectively \cite{Ortega_Kose_Canas_Chao_Unnervik_Nieto_Otaegui_Salgado_2020,Wijnands_Thompson_Nice_Aschwanden_Stevenson_2020}[19-20]

\section{Results}

Experiment I reaches a peak accuracy of 67\% before reaching a
final accuracy of 44\% with the vanilla transformer. The model is
compared to the previous state-of-the-art accuracy on NTHU-DDD,
which reached 75.4\% accuracy (Figure 3A). The training loss is
consistently lower than the test loss, indicating overfitting (Figure
3B).

\begin{figure}[ht]
  \centering
  \includegraphics[width=1\linewidth]{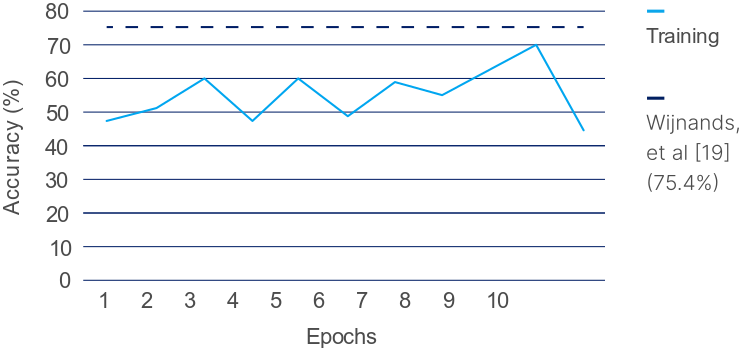}

  \caption{
  Training accuracy of the vision transformer on NTHU-DDD compared to the previous state-of-the-art with a CNN-based architecture from \cite{Wijnands_Thompson_Nice_Aschwanden_Stevenson_2020}}
  \label{fig:onecol}
\end{figure}

\begin{figure}[ht]
  \centering
\includegraphics[width=1\linewidth]{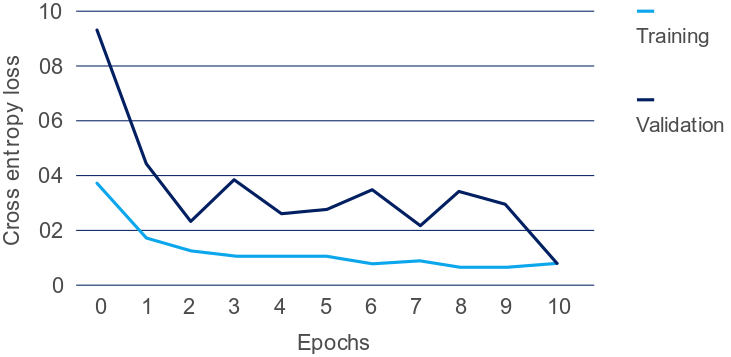}

  \caption{
  Training and validation cross-entropy loss over 10 epochs for detecting drowsy driving on NTHU-DDD with Video Swin Transformer.}
  \label{fig:onecol}
\end{figure}

Experiment II compares the test accuracy to the previous state-
of-the-art accuracy achieved with a 3D CNN from \cite{canas_dmd}. \cite{canas_dmd} reach
97.2\% accuracy while the vision transformer experiment reaches
97.5\% accuracy (Figure 4A). The test loss descends consistently
compared to training loss, indicating the transformer did not overfit.
Based on these results, a new state-of-the-art accuracy was achieved
on DMD, outperforming all previous works with 3D CNNs.

\begin{figure}[ht]
  \centering
  \includegraphics[width=1\linewidth]{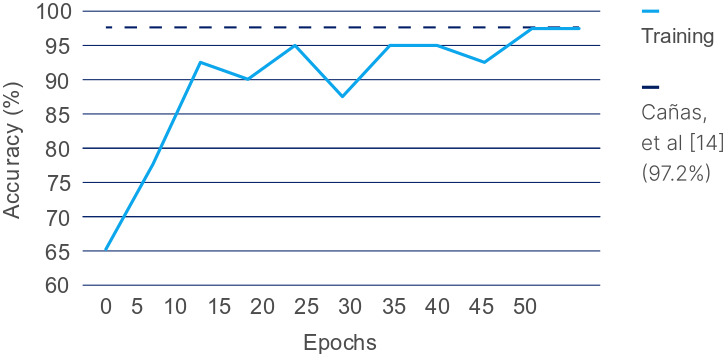}

  \caption{
Training accuracy of the Video Swin Transformer on DMD compared to the previous state-of-the-art \cite{canas_dmd} with a CNN-based architecture}  \label{fig:onecol2}
\end{figure}


\section{Conclusion}
To the best of the author’s knowledge, no experiments have been conducted applying transformers to unfit driving. In this experiment, vision transformers are evaluated for application to unfit driving as a potential solution in the future. By outperforming 3D CNNs for distracted driving, transformers are promising.

Future investigations should look to expand on the size and scope of the drowsy driving dataset to reach acceptable accuracy for drowsy driving. Future datasets should consider adding more categories to a singular, comprehensive dataset that includes drunk driving and road rage to train a more comprehensive model for unfit driving. To improve accuracy, future investigations should look to increase the number of parameters, add more data through augmentations and transformations, and use newer architectures like TokenLearner, Swin Transformer V2, and Microsoft’s Florence \cite{ryoo,Liu2022, Yuan_Chen_Chen_Codella_Dai_Gao_Hu_Huang_Li_Li_etal._2021}. Future investigations should also look to deploy the model weights to an embedded system such as a mobile device. By 2026, government regulation may mandate DMS in new cars,
and this study provides new roadways and methods to explore to
revolutionize the automotive industry and save millions of lives
\cite{Vigdor_2021}.

{\small
\bibliographystyle{ieee_fullname}
\bibliography{egbib}
}

\end{document}